\documentclass[conference]{IEEEtran} %

\newcounter{irosOrarXiv}
\usepackage{ifthen}

\ifthenelse{\value{irosOrarXiv} > 1}{
    \usepackage{xcolor}
    \usepackage[colorlinks=true,citecolor=blue!70!black,urlcolor=blue!70!black,linkcolor=blue!70!black]{hyperref}
}{
  \overrideIEEEmargins
  \usepackage{hyperref}
}

\usepackage{graphicx}

\IEEEoverridecommandlockouts 

\title{\LARGE \bf
    Natural Language as Policies:\\ Reasoning for Coordinate-Level Embodied Control with LLMs}
    
\author{Yusuke Mikami$^{1,2}$, Andrew Melnik$^{3}$, Jun Miura$^{1}$, Ville Hautamäki$^{2}$ %
\thanks{$^{1}$Department of Computer Science and Engineering at Toyohashi University of Technology, Japan.}%
\thanks{$^{2}$School of Computing at University of Eastern Finland, Finland.}%
\thanks{$^{3}$Bielefeld University, Germany.}%
\ifthenelse{\value{irosOrarXiv} > 1}{
    \thanks{Correspondence to: Yusuke Mikami \textless \href{mailto:mikami.yusuke.iv@tut.jp}{mikami.yusuke.iv@tut.jp}\textgreater}%
\thanks{Preprint. Under review.}
}{
}
}

\begin{document}

\maketitle
\thispagestyle{empty}
\pagestyle{empty}

\begin{abstract}

    We demonstrate experimental results with LLMs that address robotics task planning problems. Recently, LLMs have been applied in robotics task planning, particularly using a code generation approach that converts complex high-level instructions into mid-level policy codes. In contrast, our approach acquires text descriptions of the task and scene objects, then formulates task planning through natural language reasoning, and outputs coordinate level control commands, thus reducing the necessity for intermediate representation code as policies with pre-defined APIs. Our approach is evaluated on a multi-modal prompt simulation benchmark, demonstrating that our prompt engineering experiments with natural language reasoning significantly enhance success rates compared to its absence. Furthermore, our approach illustrates the potential for natural language descriptions to transfer robotics skills from known tasks to previously unseen tasks.
    The project website: \href{https://natural-language-as-policies.github.io}{https://natural-language-as-policies.github.io}

\end{abstract}

\IEEEpeerreviewmaketitle

\begin{figure*}[h]
    \centering
    \ifthenelse{\value{irosOrarXiv} > 1}{
        \includegraphics[width=0.95\linewidth]{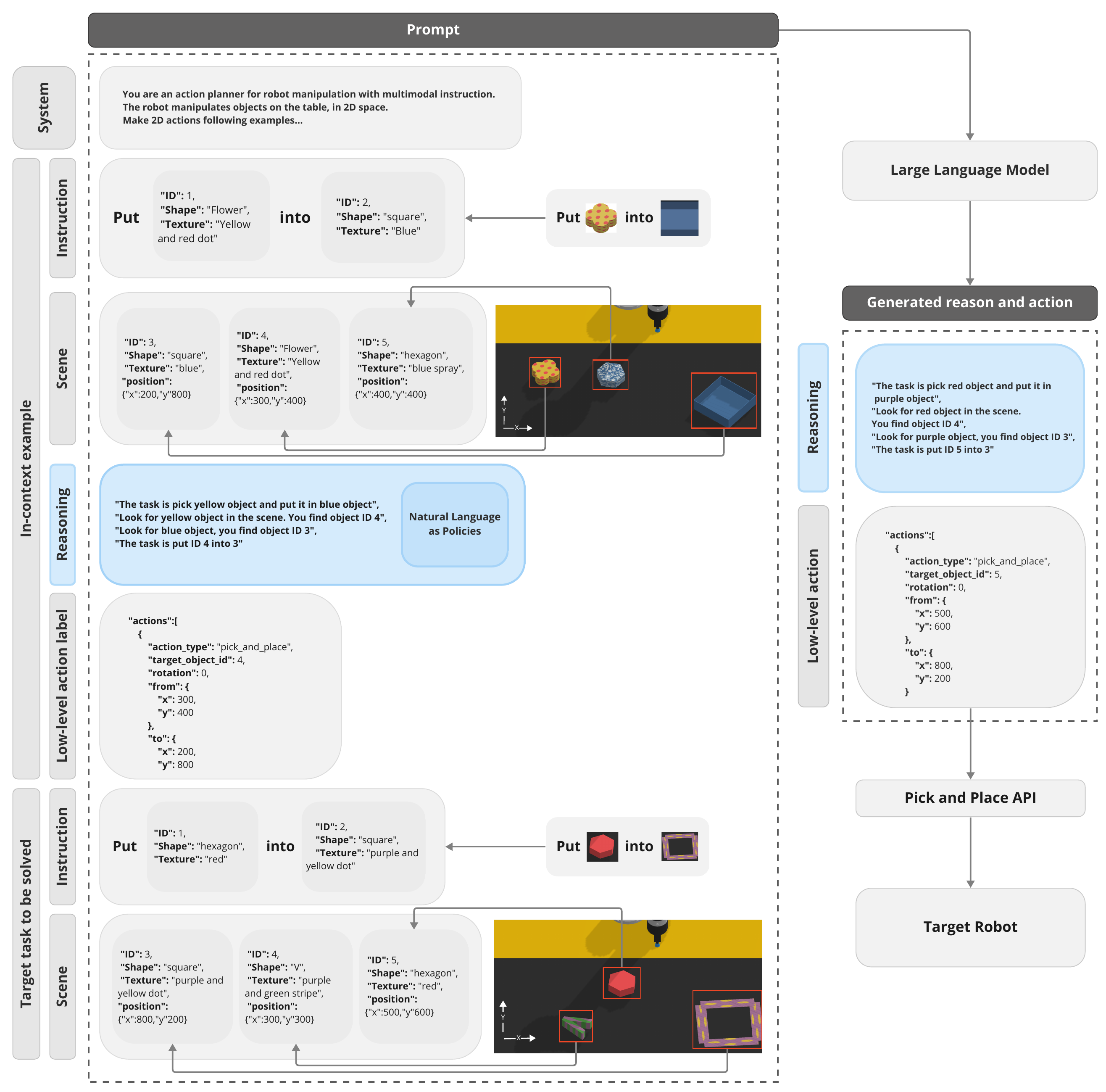}
    }{
        \includegraphics[width=0.97\linewidth]{figure/overview.pdf}
    }
    
    \caption{Overview of our approach. We provide one demonstration as an in-context example, and a planning step employing natural language reasoning instead of conventional code implementation. 
    We remove the CoT reasoning component in the in-context example for our ablation study to check the importance of natural language reasoning. 
    We use low-level API(pick-and-place or sweep) to control the robot arm. We present specific examples of natural language reasoning in Table.\ref{table:reasoning}.}
    
    \label{fig:nlap-pipeline}
\end{figure*}

\section{Introduction}

    Robotics task planning guided by human-level instruction presents a challenging topic for general robotics applications, as it entails the decomposition of high-level instructions into low-level executable robotics commands. Conventional approaches tend to address these problems in a highly task-specific and static manner and consequently struggle to achieve open-vocabulary object detection, novel task generalization, and a reduction of the training process in general.
    
    {\emph{Large language models}} (LLMs) \cite{achiam2023gpt, touvron2023llama} have had a significant impact on various applications, not only text generation tasks but also robotics task planning where LLMs attempt to interpret human-level instruction or demonstrations.
    Specifically, LLM-based robotics task planning has focused primarily on code generation approaches (CaP~\cite{liang2023code}, Progprompt~\cite{wang2023prompt}, 
    Chain of Code~\cite{li2023chain},
    SocraticModels~\cite{zeng2022socratic}, and Instruct2Act~\cite{huang2023instruct2act}) which leverage the in-context learning capability of LLM to produce code implementations by integrating predefined APIs that interface with the physical world. These studies solve the embodied control problem from an algorithmic perspective since they try to make intermediate code implementation from high-level instructions.
    
    Recently, LLM-based robotic planning has emphasized task-level zero-shot scenarios in robotic planning (Kwon et al.~\cite{kwon2023language}, Socratic Models~\cite{zeng2022socratic}) which have a huge advantage since robotics task planning often encounters novel objects, situations, and tasks. However, we propose that although LLMs can address general situations without any in-context examples, it is crucial to exploit their in-context learning capability to address a complicated novel situation and task by utilizing knowledge of previously encountered similar tasks (RAP~\cite{kagaya2024rap}).
    
    In the recent LLM-based code generation approaches, we suppose there are two primal limitations. First, code implementation itself lacks high-level contextual meaning to efficiently describe embodied skills since usually it is symbolized, indirectly connected to the scene, and abstracted in the in-context learning process. Second, these approaches are limited by task-specific pre-defined APIs, such as CLIP\
    We hypothesize that the natural language description of the whole planning process, instead of code, can contribute to removing the limitations.
    
    To overcome the limitations and advance the current state of robotics toward more semantic capability, we introduce a \emph{Chain-of-Thought} (CoT) \cite{wei2022chain}-based reasoning framework.
    In this framework, we initially possess all necessary information as text and generate the natural language reasoning and action plans without relying on pre-defined APIs.
    Our goal is to make everything explicit with natural language to efficiently describe embodied skills semantically for LLMs.
    For prompt-engineering experiments, we used tabletop manipulation tasks with multimodal prompts (VIMABench~\cite{jiang2023vima}).
    \\
    \\
    \textit{Our Contributions:}
    \begin{itemize}
        \item {Reasoning with direct interaction of environment}: Our approach enables agents to interact directly with information from the current environment.
        
        \item {Coordinate-Level action prediction}: The output of our approach consists of coordinates that can be directly executed by the target robot.
        
        \item {Teaching robots with natural language}: Our approach suggests a way for humans to teach robots to perform tasks in a manner similar to how they teach other humans.

        \item {New way to tackle novel situation}: 
        Our approach suggests a way to improve the transferability of robotics skills from a known task to a novel task at the natural language context level.
        
    \end{itemize}

\section{Related works}
    \begin{table}
        \centering
        \caption{Key components of related code generation and reasoning approaches: 
        EmbodiedGPT\cite{mu2024embodiedgpt},
        Socratic Models\cite{zeng2022socratic}, 
        Inner Monologue\cite{huang2022inner},
        Statler\cite{yoneda2023statler},
        Demo2Code\cite{wang2023demo2code}, 
        Chain of Code\cite{li2023chain},
        Progprompt\cite{singh2022progprompt},
        ChatGPT for Robotics\cite{kwon2023language},
        Code as Policies\cite{liang2023code}, 
        Instruct2Act\cite{huang2023instruct2act},
        and
        Zero-Shot Trajectory\cite{kwon2023language}.
        Our approach has differences, especially in natural language reasoning and coordinate-level output.
        }
        \label{fig:taxonomy}
        
        \includegraphics[width=1\linewidth]{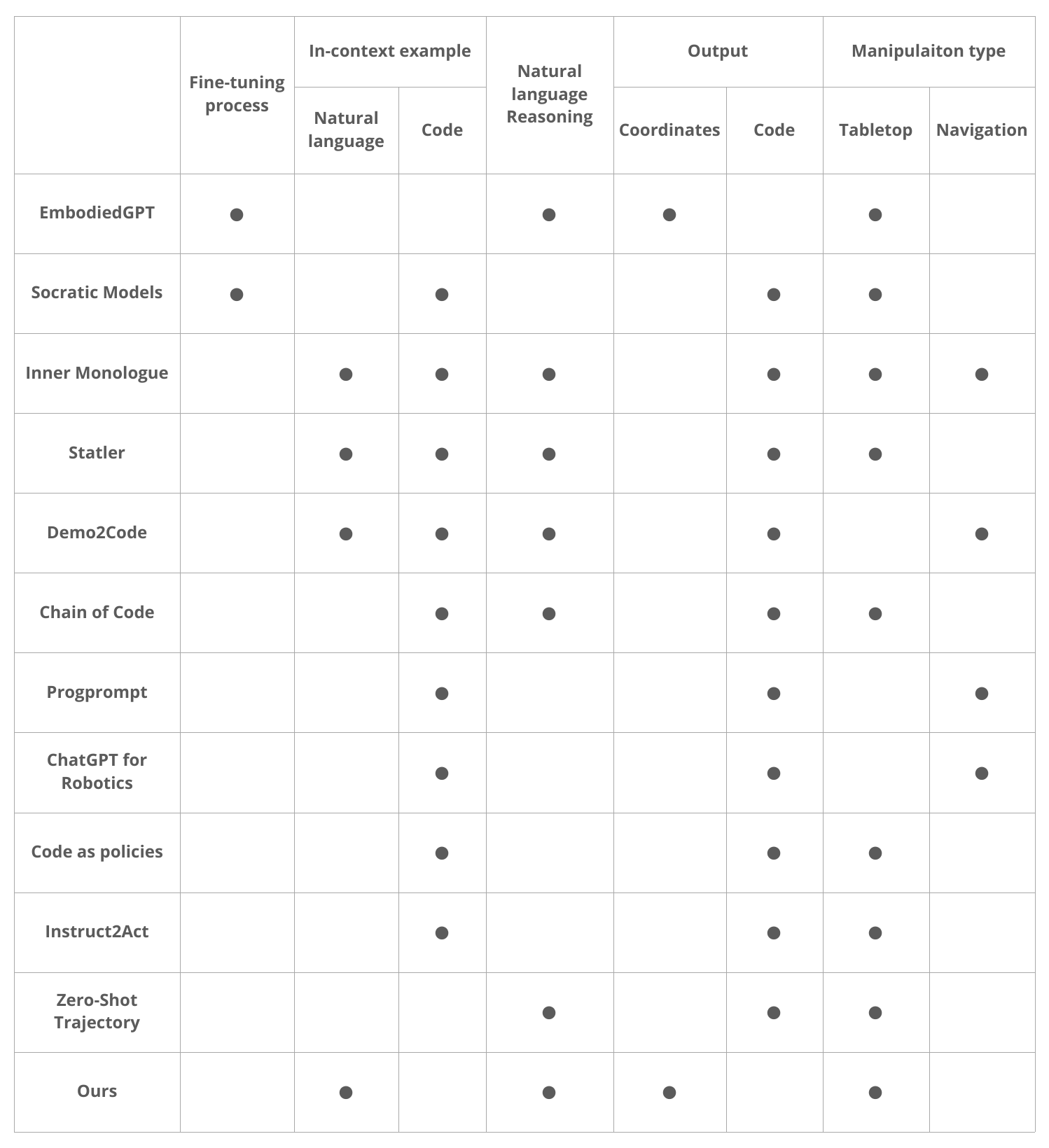}
    \end{table}

    One of the key objectives in the field of robotics is to develop a system capable of learning new tasks described in natural language, using only a handful of demonstration examples, and capable of working with an open vocabulary range of objects, similar to human abilities. Multiple attempts were made to advance existing architectures towards this long-term goal. Recent developments in Large Language and Foundation Models \cite{radford2021learning, touvron2023llama} allowed robotic architectures to make substantial progress toward this objective. 
    We present recent LLM-based approaches in Table.\ref{fig:taxonomy}.

\subsection{Imitation learning and RL for robotics task planning}
    Imitation learning and reinforcement learning (RL) \cite{bach2020learn} are common frameworks for robotics task planning. A neural network takes information from the environment, and the output comprises executable actions.
    These learning frameworks acquire robotics skills in their neural network parameters implicitly.
    These conventional approaches encounter limitations including long training duration \cite{melnik2021using}, overfitting for specific tasks \cite{schilling2019approach}, and limited input flexibility. Dataset-search policy approaches \cite{beohar2022planning, malato2024zero, malato2022behavioral, milani2023towards} propose zero-shot adaptation to provided tasks examples, thus improving in flexibility over imitation learning approach, however still having limitations for novel-task execution.
    In contrast, we leverage the reasoning capability of pre-trained LLMs with an explicit planning process to tackle the limitations. 
    
\subsection{Natural language commands to code scripts with LLMs}
    Some approaches explored the translation of natural language commands into executable code scripts \cite{chen2023robogpt, huang2023instruct2act, huang2022inner, liang2023code, Lin_2023, mu2024robocodex, singh2022progprompt, zeng2022socratic}
    where a set of examples of translation between natural language commands and executable scripts or a description of pre-defined APIs is provided, such that the model can do correct translation for new natural language commands.
    {\em Chain-of-thought} (CoT) \cite{wei2023chainofthought} enhances the reasoning abilities of large language models (LLMs) by decomposing complex tasks into smaller steps and providing examples of intermediate reasoning steps through multiple prompts. 
    Some approaches explored the integration of CoT or intermediate reasoning processes into robotics planning.
    Statler \cite{yoneda2023statler} offers a state management framework for long-horizon planning tasks.
    Demo2Code \cite{wang2023demo2code} has an efficient intermediate representation by summarizing demonstrations to produce final actions.
    Progprompt \cite{singh2022progprompt} generates situated task plans as code implementation.
    Inner Monologue \cite{huang2022inner} focuses on feedback and interaction processes in the reasoning process.
    Chain of Code \cite{li2023chain} and Language Models as Compilers~\cite{chae2024language} have advantages in both algorithmic and semantic capability by using LLM as a code interpreter.
    Text2Motion\cite{lin2023text2motion} has an intermediate symbolic and iterative process, and the outcome is a high-level command.
    On the other hand, EmbodiedGPT \cite{mu2024embodiedgpt} attempts to efficiently integrate the imitation learning process and the CoT reasoning process in its training process, however, it still requires a fine-tuning process with a customized dataset.
    In contrast to these conventional approaches, our approach solves robotics task planning by especially focusing on a semantic perspective rather than conventional algorithmic without additional fine-tuning.

\subsection{LLM-based code generation for multi-modal prompts}
    Jiang et al.\cite{jiang2023vima} introduced VIMA, which can act upon multimodal prompts within the end-to-end imitation learning approach.  Jiachen et al.\cite{li2023mastering} proposed a pre-train and fine-tune approach for the VIMA model. 
    Huang et al.\cite{huang2023instruct2act} introduced Instruct2Act which utilizes "Code as Policy"~\cite{liang2023code} to generate executable actions as a code implementation from multimodal prompts. 
    In contrast, our approach tries to achieve a flexible object detection process by describing objects and reasoning in natural language.

    \subsection{Task-level zero-shot capability}
        Task-level zero-shot capability has been a crucial problem for robotics since it involves the capability to generalization for novel tasks.
        BC-Z~\cite{jang2022bc} tackles the zero-shot problem by having a huge dataset and task embeddings. 
        Huang et al. \cite{huang2022language} introduce task planning in zero-shot situations considering executable pre-defined actions.
        Socratic Models~\cite{zeng2022socratic} leverage the zero-shot capability of LLMs to translate simple actions into code with in-context examples. 
        Teyun et al.\cite{kwon2023language} propose an LLM-based code generation approach without any in-context examples, relying on reasoning to address zero-shot tasks. 
        Obviously, these Zero-Shot solution without any examples is a promising approach, however, LLM can handle only general and simple situations and prompts in zero-shot situations \cite{rana2023contrastive}. 
        Therefore we emphasize that it is crucial to use in-context examples of previously encountered tasks if the task is complicated, for instance, VIMABench\cite{jiang2023vima}.

\subsection{Open vocabulary object detection}
    Usage of open vocabulary object detection models \cite{liu2023grounding, minderer2023scaling} is one way of embracing open vocabulary set of objects into reasoning about a task \cite{melnik2023uniteam}. Differences in objects of the same type in an environment may be complicated to express in natural language, thus integration of images of intended objects and text into multi-modal task specification \cite{tsimpoukelli2021multimodal} can provide performance benefits.
    We try to achieve open vocabulary by leveraging a pre-trained huge Vision Language model (for instance, GPT4-Vision) rather than an object detection model.

\subsection{LLMs for low-level concept}
    \begin{figure}
        \centering
        \label{fig:high-to-low}
        \includegraphics[width=1\linewidth]{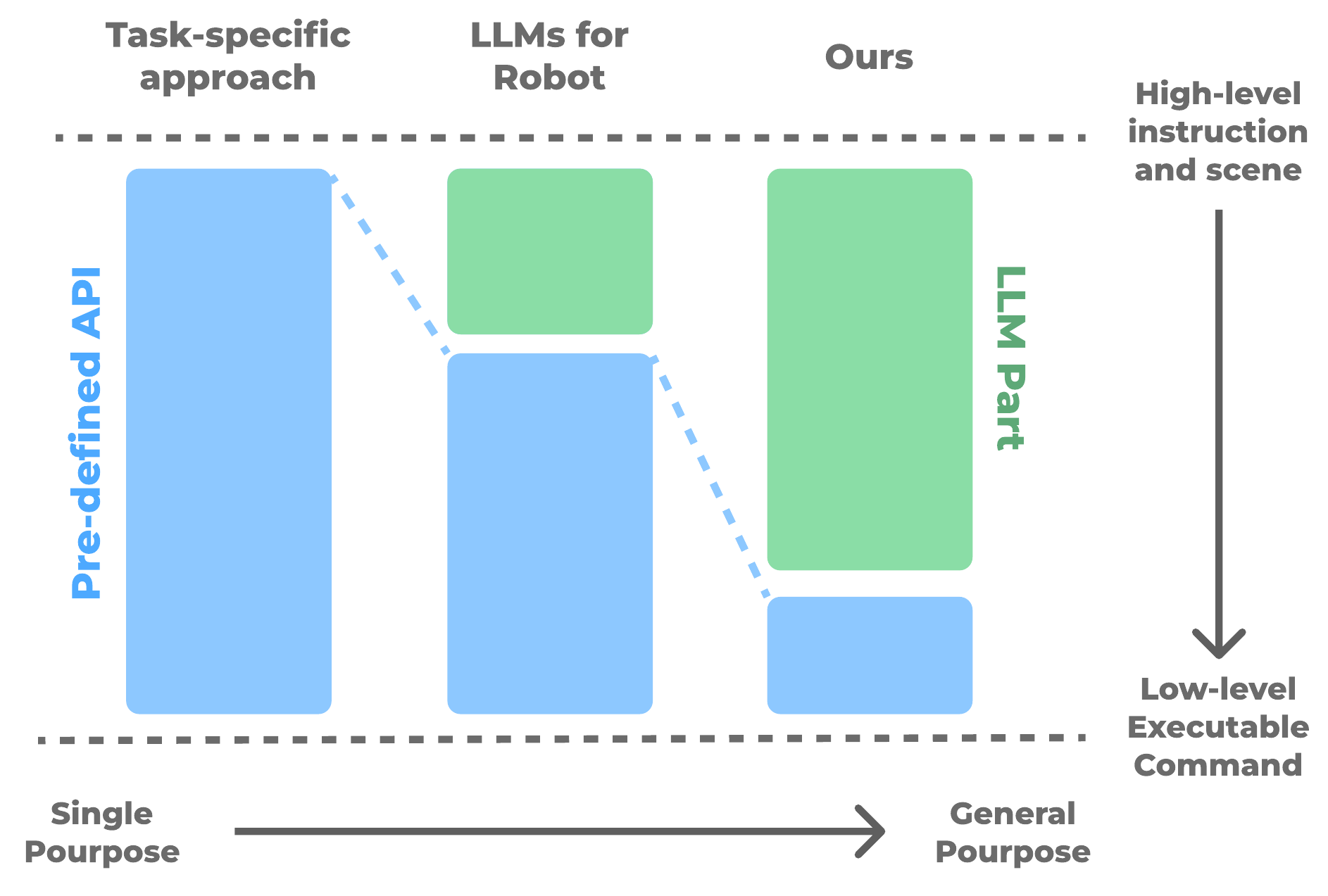}
        \caption{
        Task planning is a mapping process from high-level human intention into low-level action commands(vertical axis).
        To achieve a general-purpose agent, it is important to reduce reliance on static components.
        }
    \end{figure}

    One of the major reasons to use LLMs for agent-based systems is to acquire flexible intelligence for high-level concepts while the conventional approaches utilize static pre-defined capabilities.
    Fig.\ref{fig:high-to-low} shows how LLM can cover high-level concepts for robotics planning.
    While LLM excels at grasping high-level concepts, its proficiency at handling low-level concepts is uncertain.
    Current LLMs for OpenWorld games or robotics are utilized as an API selector or policy generator for code implementation, where pre-defined skill sets or APIs are provided
    The concept revolves around these APIs being low-level and static components, posing a significant limitation in LLM applications. 
    For instance, it cannot directly output commands like how much to turn a robot's motor or how to maneuver a Minecraft agent's body \cite{wang2023voyager}; everything occurs at a high level. This leads us to question how LLM can effectively address low-level control.  
    Several studies (\cite{tang2023saytap},\cite{wang2023prompt}) explore this field, however, they still struggle to achieve it.
    We hypothesize that it is imperative to imbue them with meaning in natural language. 
    Our approach directly outputs coordinates that have meaning in natural language reasoning, for example, we want to achieve LLM can directly produce coordinates for sweep action with the reasoning of "Sweep action starts from a position slightly away from the target object in the opposite direction to the direction you want it to move."

\section{Methods}
    While some studies focus on the code generation approach with in-context examples, we explore the decomposition of high-level tasks and the generation of coordinate-level actions with only natural language reasoning.
    
    \subsection{Problem formulation}
        Our approach solves robotic planning for tabletop manipulation. 
        The goal is to modify the state of objects in the environment to match the configuration described in the instructions.   
        Jiang et al.\cite{jiang2023vima} introduced the VIMABench framework as an open-source project for evaluating the performance of multimodal prompts. It comprises 17 tasks across four levels of generation. 

        \subsubsection{Interface of planning}
        This section provides inforamtion about the interface of VIMABench\cite{jiang2023vima} for this work.
        VIMABench provides two information for planner, which consists of multimodal prompts, which include both text and images depicting a single object and a scene with multiple objects. The scene image provided offers both top and front views, showcasing several objects within the scene.

        The output of the planning result must include specific parameters for the start and end points of the action. These parameters encompass the coordinates in the $x$ and $y$ dimensions, representing a top view for the action execution. They are utilized for both sweeping and picking actions within a two-dimensional space framework. Additionally, the rotation of the end-effector must be defined for both the start and end points. This parameter facilitates the rotation action of objects, although it is not relevant for sweep actions. During sweep actions, the rotation of the end-effector is automatically managed.
        
        \subsubsection{Actions}

        The target robot has two available actions: "Pick and place," which involves picking up an object in one location and placing it in another, and "Sweep," which entails moving objects by dragging without lifting them. These actions are selectable automatically by the benchmark depending on the task, eliminating the need for explicit action selection.

        \subsubsection{Generalization level}
            VIMABench provides four levels of generalization including placement, novel combination, novel objects, and novel task.
            Each level evaluates different zero-shot capabilities.
            In this work, we focus on only placement and novel task generalization. 
            Given that our approach does not necessitate a massive training dataset, novel combinations, and novel object generalization hold limited significance for our methodology. For novel generalization, we manually select an example from a similar task for in-context learning. 
            In VIMABench, task 10 (follow-motion) and task 13(sweep-without-touching), hold significance for novel task generalization since these tasks have novel concepts and words which is not available in other generalizations.
            Conversely, tasks 8 and 14 are deemed less crucial from the perspective of novel tasks, as they share similarities with tasks already present in the other three generalizations.
            For instance, task 8(novel-adj-and-noun) has a similar concept to tasks 6(novel-adj) and 7(novel-noun).

    \subsection{Our approach}
        
        \subsubsection{Overview}
            We introduce a robotics task planning framework to solve multimodal prompts by thinking of everything in natural language with CoT shown in Fig \ref{fig:nlap-pipeline}. 
            We first convert all images from the prompt and scene into a text description and make an action prediction with one in-context example. 
            Our approach stores robotics skills as natural language explicitly.

        \subsubsection{Pipeline}

\begin{figure}[h]
                \centering
                \includegraphics[width=1\linewidth]{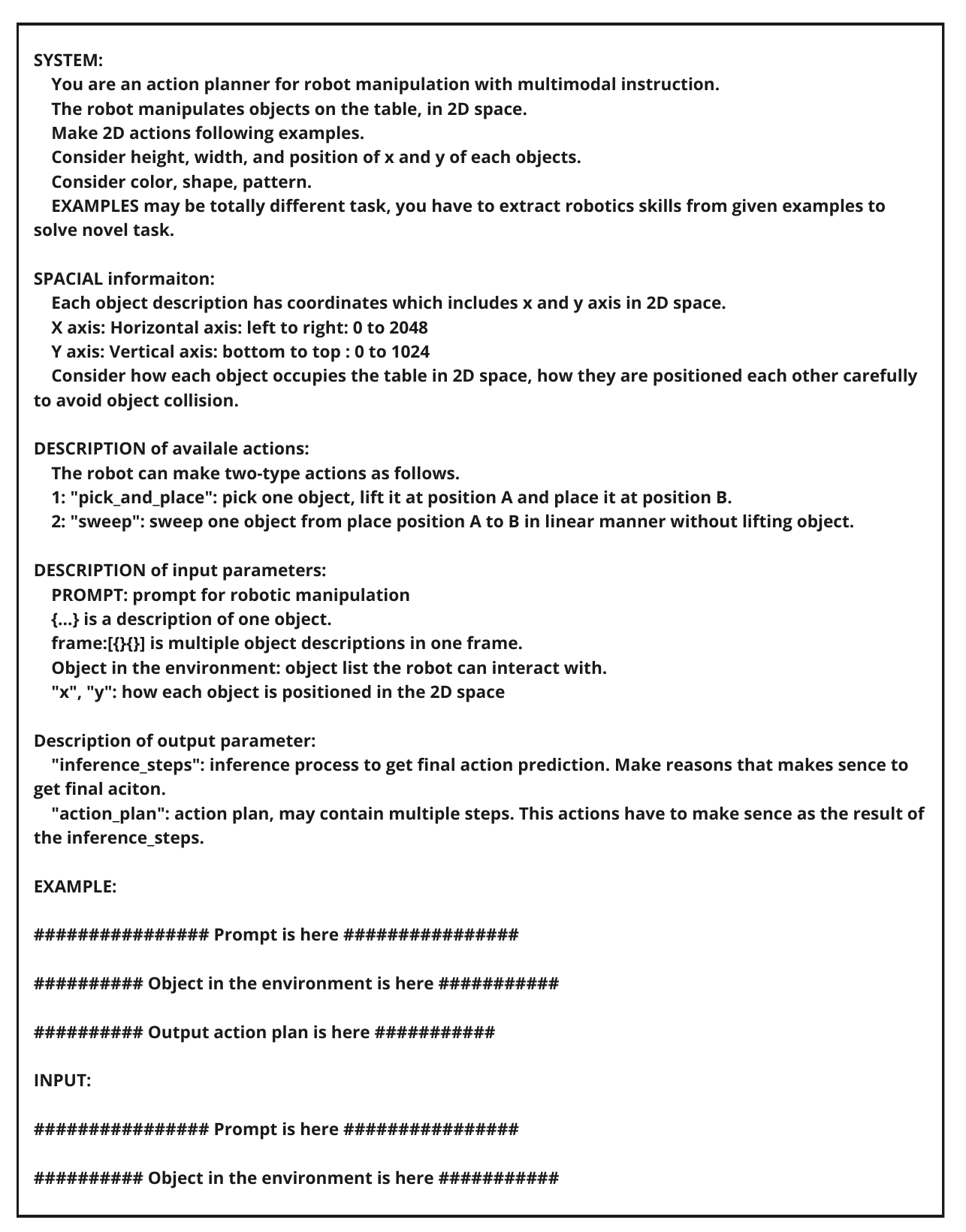}
                \caption{The full prompt with ellipses indicating omitted sections due to space limitations.}
                \label{fig:prompt}
            \end{figure}

            We translate the segmented images into text, adhering to the object description format. Subsequently, we make action predictions incorporating in-context examples with a prompt in Fig.\ref{fig:prompt}. The language model generates action predictions, including reasoning steps and actual actions, conforming to the action output format. Following this, we perform coordinate mapping, translating output coordinates from front-view to top-view. Finally, we execute the actions, which can either involve pick-and-place operations or sweeping tasks.

        \subsubsection{Object description format}
            \label{section:descirpiton-format}

            Our approach converts any images into a unified format for each object. This format includes descriptions of the shape \cite{rothgaenger2023shape} and texture of the object. 
            Additionally, it includes a section named "position," which signifies that it contains special information regarding the front view of the object. Within this section, 
            the coordinates for the center of the object are provided.

        \subsubsection{Action output format}
            \label{section:action-output-format}
            The output format comprises a continuous combination of x and y coordinates, divided into two main components. The first is the inference process, which delineates the reasoning steps leading to the final action prediction. The second component is the action plan, which outlines the steps necessary to accomplish tasks. Within the action plan, multiple plans may exist, each defined by the following parameters: action\_type (either "pick\_and\_place" or "sweep"), target\_object (the ID of the object being targeted), rotation (specifying the degree of rotation required for the target object), from (numerical values indicating the starting position of interaction), and to (indicating the ending position of interaction).

            The coordinates in the output are from the front view, then it has to be converted to the top view. We use a general mapping approach without any training process.
    
        \subsubsection{How to make reasons for each task manually}
            We generate step-by-step solutions expressed in natural language explanations for each target task independently as human-to-human teaching happens. Table \ref{table:reasoning} illustrates specific examples. Our focus primarily lies on achieving a high success rate through natural-sounding reasoning; hence, we do not overly emphasize the quality of the reasoning process. Typically, this process involves the following components: defining the target task to facilitate its decomposition, teaching object matching between the prompt and the scene, incorporating additional reasoning steps for complex tasks, and providing a specific action result as the final conclusion.

        \subsubsection{LLMs}
            We employ GPT-3.5 as our primary experimental framework, supplemented by GPT-4 for additional experimentation. Utilizing their respective assistant APIs facilitates the efficient provision of system prompts and inputs. 

        \subsubsection{Limitations}
             We remove task 9 "twist" and task 8 "novel\_adj\_and\_noun" from our experiment due to the limitation of our approach. For instance, our approach cannot detect an exact rotation of objects for task 9(twist) task.

    \subsection{Ablation study}
        Our approach focuses on the importance of reasoning with natural language. Therefore, the ablation study is critical. As described in Fig.\ref{fig:nlap-pipeline}, we remove the reasoning part from our framework.
        In Table.\ref{fig:table_rcot_result}, the "Without Chain of Thought" (w/o CoT) condition indicates that the input of the Language Model lacks a manual CoT example, yet the output of the LLM includes CoT reasoning steps. This condition operates as a zero-shot reasoning scenario.

    \subsection{Result}

        Table.\ref{fig:table_rcot_result} and \ref{fig:table_rcot_result_2} show the success rate as a table for different generalizations.
        Each of our models and tasks undergoes at least 30 attempts. 
        Table.\ref{table:gpt4} shows an additional experiment with GPT4 only for task 10, the follow-motion task. This additional experiment has 10 attempts only for our approach.

\begin{table*}
    \centering
    \caption{
    Placement generalization: Success rate comparison of our approach with existing approaches through ablation studies. 
    The bold numbers indicate the average value. 
    We compare our approach with VIMA (200M) \cite{jiang2023vima}, and Instruct2Act \cite{huang2023instruct2act}.  
    Our approach utilizes the model GPT-3.5-Turbo-1106 from OpenAI and is described with ablation studies (with CoT and without CoT input). "w/o CoT" indicates an ablation study as described in Fig.\ref{fig:nlap-pipeline}. 
    The "one-shot-example-task" column indicates which task is used as an in-context example for our approach. 
    Each of our models and tasks undergoes at least 30 attempts. 
    We remove task 9 ("twist") due to the limitation of our approach. 
    }
        
    \begin{tabular}{ccccccc}
    \hline
        Task Num & Task & One-shot example for Ours & VIMA 200M & Instruct2Act & Ours (w/o CoT) & Ours \\ \hline
        1 & visual\_manipulation & visual\_manipulation & 100 & 91 & 93 & 100 \\
        2 & scene\_understanding & scene\_understanding & 100 & 81 & 60 & 67 \\
        3 & rotate & rotate & 100 & 98 & 93 & 93 \\
        4 & rearrange & rearrange & 100 & 79 & 52 & 73 \\
        5 & rearrange\_then\_restore & rearrange\_then\_restore & 57 & 72 & 25 & 73 \\
        6 & novel\_adj & novel\_adj & 100 & 82 & 13 & 43 \\
        7 & novel\_noun & novel\_noun & 100 & 88 & 8 & 80 \\
        11 & follow\_order & follow\_order & 77 & 72 & 0 & 0 \\
        12 & sweep\_without\_exceeding & sweep\_without\_exceeding & 93 & 68 & 17 & 47 \\
        15 & same\_shape & same\_shape & 97 & 78 & 10 & 80 \\
        16 & manipulate\_old\_neighbor & manipulate\_old\_neighbor & 77 & 64 & 8 & 20 \\
        17 & pick\_in\_order\_then\_restore & pick\_in\_order\_then\_restore & 43 & 85 & 10 & 30 \\ \hline
         &  &  & \textbf{87} & \textbf{80} & \textbf{32} & \textbf{59} \\ \hline
    \end{tabular}%
    \label{fig:table_rcot_result}
    
\end{table*}

\begin{table*}
\centering
    \caption{Novel task generalization: Success rate comparison of our approach with existing approaches through ablation studies. 
    The bold numbers indicate the average value. We compare our approach with VIMA (200M) \cite{jiang2023vima}, and Instruct2Act \cite{huang2023instruct2act}.  Our approach utilizes the model GPT-3.5-Turbo-1106 from OpenAI and is described with ablation studies (with CoT and without CoT input). "w/o CoT" indicates an ablation study as described in Fig.\ref{fig:nlap-pipeline}. The "one-shot-example-task" column indicates which task is used as an in-context example for our approach. Each of our models and tasks undergoes at least 30 attempts. 
    We remove task 8 ("novel\_adj\_and\_noun") due to the limitation of our approach.
    }

\resizebox{\textwidth}{!}{%
\begin{tabular}{ccccccc}
\hline
Task Num & Task & One-shot example for Ours & VIMA 200M & Instruct2Act & Ours (w/o CoT) & Ours \\ \hline
10 & follow\_motion & rearrange\_then\_restore & 0 & 35 & 0 & 12 \\
13 & sweep\_without\_touching & sweep\_without\_exceeding & 0 & 0 & 0 & 3 \\
14 & same\_texture & same\_shape & 95 & 80 & 3 & 71 \\ \hline
 &  &  & \textbf{32} & \textbf{38} & \textbf{1} & \textbf{29} \\ \hline
\end{tabular}%
}
\label{fig:table_rcot_result_2}
\end{table*}

\begin{table*}
    \centering
    \caption{
    Novel task generalization: Success rate comparison for additional experiment with GPT4 only for task 10, follow-motion task. 
    }
    \resizebox{\textwidth}{!}{
    \begin{tabular}{ccccccc}
    \hline
    Task Num & Task & One-shot example for Ours & VIMA 200M & Instruct2Act & Ours (GPT3.5) & Ours (GPT4) \\ \hline
    10 & follow\_motion & rearrange\_then\_restore & 0 & 35 & 12 & 90 \\ \hline
    \end{tabular}
    }
    \label{table:gpt4}
\end{table*}

\section{Discussion}
    Overall, our quantitative results demonstrate that the natural language reasoning process has a critical role in performing a better success rate, especially for novel-task generalization. On the other hand, our approach does not perform better than other existing approaches in most tasks.
    
    \textit{Importance of Chain of Thought reasoning:}
        According to Table. \ref{fig:table_rcot_result} and \ref{fig:table_rcot_result_2}, Chain-of-Thought reasoning is crucial for almost all tasks as our hypothesis mentions. 
        The success rate improves from 32\% to 59\% in placement generalization (Table.\ref{fig:table_rcot_result}) with CoT reasoning.
        This result demonstrates that a natural language reasoning step is a critical component in producing low-level action prediction, especially for novel task generalization (Table.\ref{fig:table_rcot_result_2}).
        
    \textit{Skill extraction for novel task:}
        Our results suggest that the explicitness of whole robotics task planning can have the potential to transfer robotics skills from known to novel tasks with LLMs.
        To tackle the novel-task generalization of VIMABench, it is required to extract essential skills from in-context examples. Our ablation study suggests that the natural language reasoning process efficiently contributes to novel tasks. Especially for the follow-motion task, GPT4 performs 90\% in extracting skills from the rearrange-and-restore task to solve the follow-motion task as described in Table.\ref{table:gpt4}.
        On the other hand, our approach still struggles to solve task 13 (sweep-without-touching) which requires understanding a novel concept of collision avoidance.

    \textit{Numerical values for LLMs:}
        We demonstrate that LLM can effectively handle numerical values for most tasks. Although both the input and output of our approach include numerical values, LLM successfully generates action plans containing coordinates. However, our approach still struggles to solve task 11, a follow-order task that requires how objects are stacked based on their coordinates.
        Additionally, We show that LLM has spatial understanding when the coordinates of each object are injected as an input.  Our result of task 12(sweep-without-exceeding), suggests that LLM can handle numerical values of actions that can not be described with object ID.
        CaP \cite{liang2023code} mentioned that code-based reasoning outperforms natural language CoT for spatial-geometric reasoning, however, our study provides a novel grounding approach for LLMs not only for geometric reasoning but also for planning itself.

    \textit{Multi-steps action prediction:} 
        Since our approach does not predict actions in an autoregressive manner, all actions must be predicted simultaneously. 
        Task 5 (Table.\ref{fig:table_rcot_result}) suggests that the CoT reasoning process can support long-step action prediction effectively.
        Tasks 4 and 5 (Table.\ref{fig:table_rcot_result}) require multi-step actions where the target object's location changes in each step.

    \textit{Open problems:}
        We have identified several challenges in our current approach. 
        Firstly, there is the limited capability of text description capability of objects. Extracting nuanced information such as object height, rotation, or darkness solely using a Vision-Language model, rather than a deterministic approach, is particularly difficult.
        Secondly, there's the matter of feedback. The current Language-Logic Model (LLM) operates a robot within a closed loop, lacking the capacity to integrate feedback from its performance. 
        Thirdly, our approach is limited in its ability to generate complex actions, currently only capable of producing tabletop actions. 
        Fourthly, our approach focuses on semantic capability for planning problems, whereas robotic task planning typically necessitates algorithmic capability as well.

\section{Conclusion}

    We introduce an LLM-based concept wherein planning occurs solely within a natural language framework, conferring advantages over conventional LLM-based code-generation methodologies.
    Our quantitative results demonstrate that our approach does not consistently outperform other existing approaches across various tasks, however, it shows considerable potential.
    Importantly, our approach underscores the capability to tackle novel tasks with known skill sets. 
    In future endeavors, we envision applying our approach to diverse tasks and situations leveraging the flexible reasoning capability of our approach and investigating novel task generalization.

\section*{APPENDIX}

Table.\ref{table:reasoning} shows a natural language reasoning we manually made for in-context examples.

\begin{table*}[h]
    \centering
    
    \caption{Specific example of natural language reasoning we manually made for in-context learning. We do not have any specific format to produce these reasonings and we try to make natural reasoning humanly.}
    
    \includegraphics[width=1\linewidth]{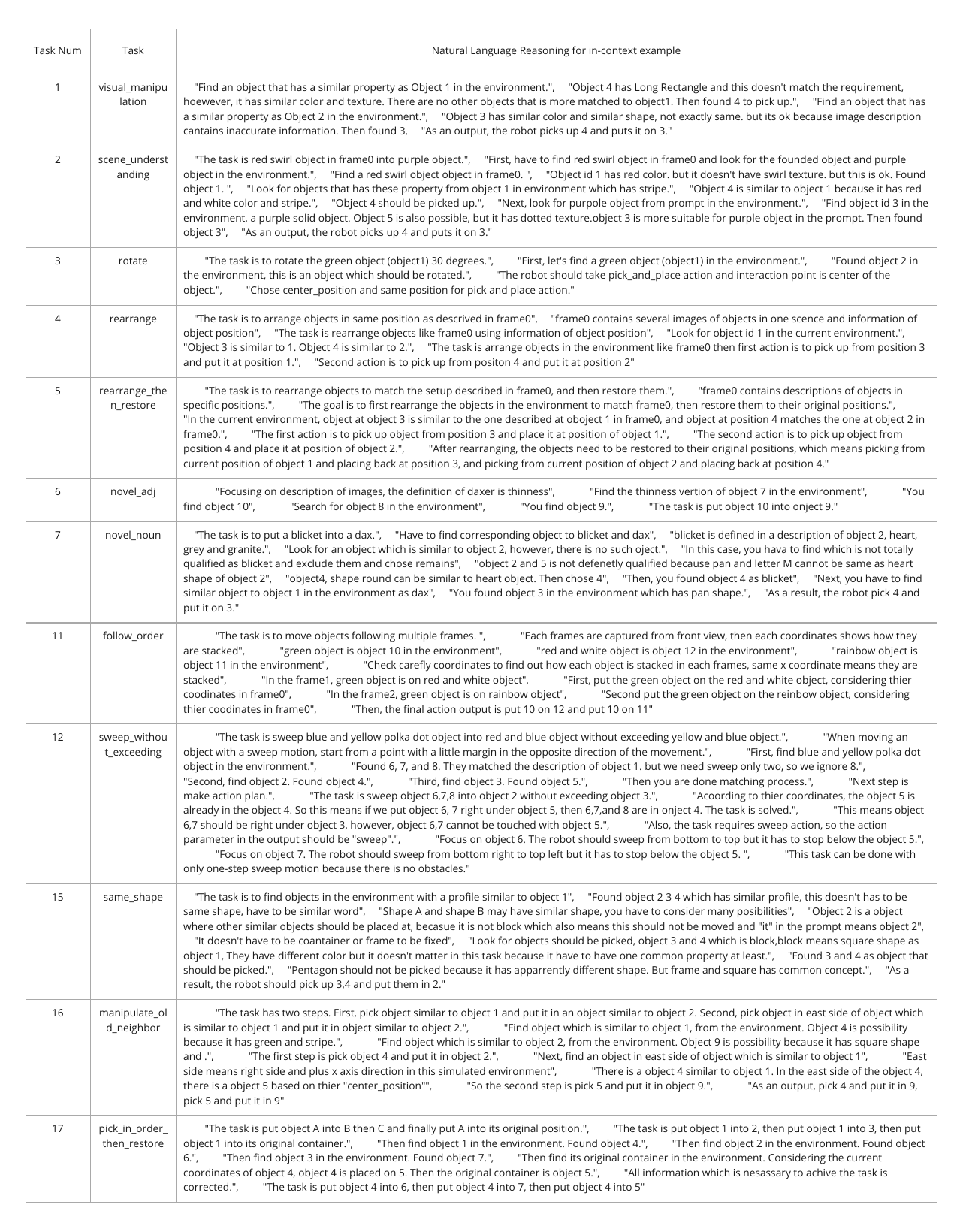}
    
    \label{table:reasoning}
\end{table*}

\bibliography{references}

\end{document}